\documentclass[10pt,conference]{IEEEtran}
\IEEEoverridecommandlockouts
\usepackage{cite}
\usepackage{amsmath,amssymb,amsfonts}
\usepackage{algorithmic}
\usepackage{graphicx}
\usepackage{textcomp}

\usepackage{xcolor}
\usepackage{indentfirst} 
\usepackage{setspace}
\usepackage{geometry}
\usepackage{graphicx}
\usepackage{booktabs}
\usepackage{threeparttable}
\usepackage{caption}
\usepackage{placeins}
\usepackage{listings}
\usepackage{xcolor}

\usepackage[ruled,vlined]{algorithm2e}
\usepackage{url}
\def\BibTeX{{\rm B\kern-.05em{\sc i\kern-.025em b}\kern-.08em
    T\kern-.1667em\lower.7ex\hbox{E}\kern-.125emX}}
\begin{document}

\title{Semi-Supervised Mixture Models under the Concept of Missing at Radom with Margin Confidence and Aranda–Ordaz Function\\}

\author{\IEEEauthorblockN{1\textsuperscript{st} Jinyang Liao}
\IEEEauthorblockA{\textit{School of Mathematics and Statistics} \\
\textit{University of New South Wales}\\
Sydney, Australia \\
jinyang.liao@student.unsw.edu.au}
\and
\IEEEauthorblockN{2\textsuperscript{nd} Ziyang Lyu}
\IEEEauthorblockA{\textit{School of Mathematics and Statistics} \\
\textit{University of New South Wales}\\
Sydney, Australia \\
ziyang.lyu@unsw.edu.au}
}

\maketitle

\begin{abstract}
This paper presents a semi-supervised learning framework for Gaussian mixture modelling under a Missing at Random (MAR) mechanism. The method explicitly parameterizes the missing-label mechanism by modelling the probability of missingness as a function of classification uncertainty. To quantify classification uncertainty, we introduce margin confidence and incorporate the Aranda–Ordaz (AO) link function to flexibly capture the asymmetric relationships between uncertainty and missing probability. Based on this formulation, we develop an efficient Expectation–Conditional Maximization (ECM) algorithm that jointly estimates all parameters appearing in both the Gaussian mixture model (GMM) and the missingness mechanism, and subsequently imputes the missing labels by a Bayesian classifier derived from the fitted mixture model. This method effectively alleviates the bias induced by ignoring the missingness mechanism while enhancing the robustness of semi-supervised learning. The resulting uncertainty-aware framework delivers reliable classification performance in realistic MAR scenarios with substantial proportions of missing labels.

\end{abstract}


\section{Introduction}
Modern machine learning applications often contend with partially labeled datasets, in which labels are available for merely a small proportion of the data, while the majority remain unlabeled. According to \cite{VanEngelen2020,Zhu2005,Wu2017,MartinezUso2010}, this scenario generally arises when labeling is expensive or infeasible for all data points. To provide a principled framework for reasoning about such incomplete data, \cite{Rubin1976} introduced the widely used classification of missing mechanisms into Missing Completely at Random (MCAR), Missing at Random (MAR), and Missing Not at Random (MNAR). Subsequently, \cite{Little1986,Heitjan1996} clarified Rubin’s taxonomy by emphasizing the conceptual and inferential distinction between MCAR and MAR, highlighting the critical challenge that labels in practice are typically not MCAR. In practice, missingness can be characterized within the MAR framework by \cite{Bennett2001,Yan2015,Mealli2015}, which allows missingness to depend on observed characteristics such as classification uncertainty, but not on unobserved labels. This property has substantial implications, as evidence from \cite{Ahfock2020,Nakagawa2015,Santos2019} indicates that overlooking this structure can introduce bias in parameter estimation and weaken classification performance in both discrimination and calibration. 
Accordingly, explicitly modeling the missing-label mechanism is essential for accurate estimation and robust predictive performance in semi-supervised settings. Under the MAR assumption, \cite{Lyu2024,lyu2024analysis} incorporates the dependence of missingness on observable characteristics.
Building on the finite mixture modeling framework of \cite{McLachlan2000} and following \cite{Lyu2024, McLachlan2019} together with the semi-supervised learning treatment of \cite{Ahfock2020}, this article focuses on MAR within a two-component Gaussian mixture modeling framework and develops a new specification of MAR within the finite mixture setting.

In modeling the MAR missingness mechanism, we introduce a new concept of margin confidence to quantify the classification uncertainty of each observation\cite{Zhang2010,Ibrahim2002, Wang2020}. Compared with previous work that relies on Shannon entropy\cite{Lyu2024, McLachlan2019}, margin confidence provides a substantially more efficient approximation measure while retaining strong discriminative power. We further show that Shannon entropy is well approximated by the squared margin confidence in most practical settings, enabling significant computational savings with negligible loss in model performance.
In addition, previous approaches commonly use logit link to model MAR mechanism. However, in complex settings such as non-Gaussian mixtures, the logit link exhibits limited robustness, as it enforces a symmetric response around the decision boundary and is therefore less flexible in capturing highly skewed or uncertainty-concentrated missingness mechanisms. In contrast, the Aranda–Ordaz link generalizes the logit by introducing a shape parameter $\lambda$, which relaxes the inherent symmetry of the logistic function\cite{ArandaOrdaz1981,Scalera2021RobustLink}. By adjusting $\lambda$, the link allows the missing probability to depend asymmetrically on margin-based uncertainty \cite{Scalera2021RobustLink,Pumi2020AO,Dehbi2016AOQR}. This choice markedly improves the stability and adaptability of the proposed MAR-aware mixture model.

In the experimental section, we evaluate the proposed framework using both controlled simulations and real-data experiments. 
In the simulation section, labels are removed by a margin-based MAR mechanism, allowing comparison between the proposed ECM--AO and a logistic baseline that ignores the missingness mechanism. 
To assess robustness beyond idealized settings, we further consider non-Gaussian mixture models and the MAGIC Gamma Telescope dataset. These experiments examine the performance of ECM--AO under distributional misspecification and realistic label sparsity.

\section{Methodology}\label{MM}

\subsection{Gaussian mixture model}

Let $\boldsymbol{Y}$ be a random vector of $d$ dimensional arising from a $K$-component Gaussian mixture model.
Each component $k$ $(k=1,\ldots,K)$ is associated with a prior probability $\pi_k$, where $\pi_k \geq 0$ and $\sum_{k=1}^{K}\pi_k = 1$. 
Conditional on assignment to component $k$, the distribution of $\boldsymbol{Y}$ is multivariate Gaussian with density $\mathcal{N}(\boldsymbol{y};\boldsymbol{\mu}_k,\boldsymbol{\Sigma}_k)$,
where $\boldsymbol{\mu}_k \in \mathbb{R}^d$ denotes the mean vector and $\boldsymbol{\Sigma}_k \in \mathbb{R}^{d\times d}$ denotes the covariance matrix of the $k$-th component. 
The resulting marginal density of $\boldsymbol{Y}$ is therefore
\[
f(\boldsymbol{y}_j;\boldsymbol{\theta}) = \sum_{k=1}^{K} \pi_k \, \mathcal{N}(\boldsymbol{y}_j;\boldsymbol{\mu}_k,\boldsymbol{\Sigma}_k),
\]
with the complete parameter vector given by
\[
\boldsymbol{\theta} = \big\{\pi_1,\ldots,\pi_{K-1},\, \boldsymbol{\mu}_1,\ldots,\boldsymbol{\mu}_K,\, \boldsymbol{\Sigma}_1,\ldots,\boldsymbol{\Sigma}_K \big\}.
\]

For an observation $\boldsymbol y_j$, the Bayes' rule of allocation assigns it to class
$z_j$ if
$z_j = \arg\max_{k \in \{1,\ldots,K\}} \tau_{kj}$,
where 
$
\label{eq:responsibility}
\tau_{kj}
= {\pi_k \, \mathcal{N}\!\left(\boldsymbol{y}_j\mid \boldsymbol{\mu}_k,\boldsymbol{\Sigma}_k\right)}/
       {\sum_{j=1}^{K} \pi_j \, \mathcal{N}\!\left(\boldsymbol{y}_j\mid \boldsymbol{\mu}_j,\boldsymbol{\Sigma}_j\right)}
$
denotes the posterior probability that observation $\boldsymbol y_j$ belongs to the class
$k$, given $\boldsymbol Y = \boldsymbol y_j$.

Consider a training set $\{\boldsymbol y_j, m_j, z_j\}_{j=1}^n$, where $\boldsymbol y_j \in \mathbb R^d$ denotes the feature vector, $z_j \in \{1,\ldots,K\}$ is the class label, and $m_j$ is the missingness indicator taking the value $m_j=0$ if $z_j$ is observed and $m_j=1$ otherwise.

To facilitate the formulation of the likelihood, we define the indicator variable $z_{ij}$ such that $z_{ij}=1$ if observation $\boldsymbol y_j$ belongs to component $i$ and $z_{ij}=0$ otherwise, for $i=1,\ldots,K$ and $j=1,\ldots,n$. 
Following the framework of \cite{Lyu2024}, the observed data log-likelihood can be decomposed into contributions from labeled and unlabeled samples given by
\begin{align}
&\log L_{C}(\boldsymbol\theta) = \nonumber \\
&\sum_{j=1}^n (1-m_j)\sum_{i=1}^K z_{ij}\,
   \log\Big\{\pi_i \mathcal N(\boldsymbol y_j;\boldsymbol\mu_i,\boldsymbol\Sigma_i)\Big\}, 
   \label{eq:LC}\\[4pt]
&\log L_{UC}(\boldsymbol\theta) = 
\sum_{j=1}^n m_j \log\Bigg(
    \sum_{i=1}^K \pi_i \mathcal N(\boldsymbol y_j;
\boldsymbol\mu_i,\boldsymbol\Sigma_i)\Bigg).
   \label{eq:LUC}
\end{align}
The total observed-data log-likelihood is then
\begin{equation}
\log L_{PC}^{(\text{ig})}(\boldsymbol\theta) 
= \log L_{C}(\boldsymbol\theta) + \log L_{UC}(\boldsymbol\theta).
\label{eq:LPCzij}
\end{equation}

\subsection{Margin Confidence}

Margin confidence quantifies the model's certainty in its prediction by measuring the difference between the highest and the second-highest posterior class probabilities for a given input. For a data point $\boldsymbol{y}_j$, the margin confidence $\delta_j$ is defined as
\begin{equation*}
\delta_j = \tau_{(1)j} - \tau_{(2)j},
\end{equation*}
where $\tau_{(1)j}$ and $\tau_{(2)j}$ denote the largest and second largest posterior probabilities.

In the case of binary classification, the margin confidence admits the following simplified form:
\begin{equation}
\delta_j = \left| 2\tau_{1j} - 1 \right|,
\end{equation}
where $\tau_{1j}$ is the posterior probability that $y_j$ belongs to class $1$. A margin confidence value close to $1$ indicates a confident classification, while values near $0$ correspond to high predictive uncertainty.

Based on the derivation in the supplementary material(Section~S1), the relationship between margin confidence and entropy can be expressed as follows:

\begin{equation}
H(\tau) \approx \log 2 - \frac{\delta_j^2}{2},
\end{equation}
which reveals a linear relationship between entropy and squared margin confidence.

\subsection{Missing at random mechanism}
Following the framework of \cite{Rubin1976}, we assume that the missingness is MAR, which means it depends only on the the observed feature vector. In this context, the missingness mechanism is specified as

\[
\begin{aligned}
\Pr(M_j = 1 \mid y_j, z_j)&=\Pr(M_j = 1 \mid y_j)
\\&=q(y_j;\boldsymbol{\theta}, \boldsymbol{\xi}),
\end{aligned}
\]

where $M_j \in \{0,1\}$ denotes the missingness indicator for observation $j$, with $M_j = 1$ indicating a missing label and $M_j = 0$ otherwise, and \( \boldsymbol{\xi} = (\xi_0, \xi_1) \) parameterizes the missingness mechanism. 

Following \cite{Ahfock2020}, missing probability is modeled as a logit function of log--entropy \( \log H(y_j; \theta) \).
Based on the derivation in the supplementary material(Section~S2), \( \log H(y_j; \theta) \) is shown to have an approximately linear relationship with \( \delta_j^2 \). This yields a new model for the missing probability:

\[
    q(y_j; \boldsymbol{\theta},\boldsymbol{\xi}) = \frac{\exp\left(\xi_0 + \xi_1 \delta_j^2\right)}{1 + \exp\left(\xi_0 + \xi_1 \delta_j^2\right)}.
\]

\subsection{Aranda--Ordaz Link Function}

Based on \cite{ArandaOrdaz1981}, the AO link generalizes the logit link by allowing asymmetric covariate effects on the missing probability and is defined as

\[\label{eq:ao-link}
g(q;\lambda) \;=\;
\begin{cases}
\displaystyle \log \!\left( \dfrac{(1-q)^{-\lambda} - 1}{\lambda} \right) & \lambda \neq 0, \\[1.2em]
\displaystyle \log \!\big(-\log(1-q)\big) & \lambda = 0 ,
\end{cases}
\]
where $q$ is the missing probability and $\lambda$ is a shape parameter that controls the degree of asymmetry. 

Under the MAR assumption, we model the missing probability by squared margin confidence. Thus, $q$ can be written as $q(\delta_j;\boldsymbol{\alpha},\lambda)$, where $\boldsymbol{\alpha}=(\alpha_0, \alpha_1)$ parameterizes the missingness mechanism. The corresponding inverse link takes the form
\begin{gather*}
q(\delta_j;\boldsymbol{\alpha},\lambda)
= g^{-1}(\eta;\boldsymbol{\alpha},\lambda)
= 1 - \bigl(1 + \lambda e^\eta \bigr)^{-1/\lambda}, \\
\text{where } \eta = \alpha_0 + \alpha_1 \delta_j^2 .
\end{gather*}

In practical applications, the parameter $\lambda$ can be predetermined. Alternatively, if higher modeling accuracy is desired, $\lambda$ can be incorporated into the ECM algorithm and updated iteratively.

We define the complete parameter set as $\boldsymbol{\Theta} = \{\lambda, \boldsymbol{\theta}, \boldsymbol{\alpha}\}$. Substituting into the above expression yields the result as 
\begin{equation}\label{eq:ao}
q(\boldsymbol{y}_j;\boldsymbol{\Theta})= 1 - \bigl(1 + \lambda e^{\alpha_0 + \alpha_1 \delta_j^2} \bigr)^{-1/\lambda}.
\end{equation}

\subsection{log likelihood function under missing at random mechanism}

In order to incorporate the MAR mechanism into the likelihood of the Gaussian mixture model, the density function of the training set is divided into $f(\boldsymbol{y}_j,z_j,m_j=0;\boldsymbol{\Theta})$ and $f(\boldsymbol{y}_j,m_j=1;\boldsymbol{\Theta})$ by the missingness indicator $m_j$. Conditional on $m_j=0$,
\begin{align*}
    &f(\boldsymbol y_j, z_j, m_j=0;\boldsymbol{\Theta}) = \\&\prod_{i=1}^K \Big\{ \pi_i \mathcal N(\boldsymbol y_j;\boldsymbol\mu_i,\boldsymbol\Sigma_i) \Big\}^{z_{ij}} \, \Big\{1 - q(\boldsymbol y_j;\boldsymbol\Theta)\Big\}.
\end{align*}
Conditional on $m_j=1$,
\[
    f(\boldsymbol{y}_j,m_j=1;\boldsymbol{\Theta}) = \sum_{i=1}^K \pi_i \mathcal N(\boldsymbol y_j;\boldsymbol\mu_i,\boldsymbol\Sigma_i)\, q(\boldsymbol y_j;\boldsymbol\Theta).
\]
Based on these two different density functions, the full likelihood function for $\boldsymbol\Theta$ is given by 
\begin{align*}
  &L_{PC}^{(\text{full})}(\boldsymbol\Theta)=
  \prod_{j=1}^n 
\Big\{ f(\boldsymbol y_j, z_j, m_j=0;\boldsymbol\Theta) \Big\}^{\,1-m_j}
\\&\Big\{ f(\boldsymbol y_j, m_j=1;\boldsymbol\Theta) \Big\}^{\,m_j}.
\end{align*}
In recent work, it has been shown that the full log-likelihood function for $\boldsymbol\Theta$ can be expressed as
\begin{equation}
\log L_{PC}^{(\text{full})}(\boldsymbol\Theta) 
= \log L_{PC}^{(\text{ig})}(\boldsymbol\theta) 
+ \log L_{PC}^{(\text{miss})}(\boldsymbol\Theta),
\end{equation}
where
\begin{align}
&\log L_{PC}^{(\text{miss})}(\boldsymbol\Theta)= \nonumber
\sum_{j=1}^n \Big[ (1-m_j)\,\log\{1-q(\boldsymbol y_j;\boldsymbol\Theta)\} 
\\&+ m_j \,\log q(\boldsymbol y_j;\boldsymbol\Theta) \Big],
\end{align}
and the terms $\log L_{PC}^{(\text{ig})}(\boldsymbol\theta)$ and $q(\boldsymbol y_j;\boldsymbol\Theta)$ are formally specified in equations~\ref{eq:LPCzij} and~\ref{eq:ao}\cite{Lyu2024}.

To obtain the maximum-likelihood estimates of $\boldsymbol\Theta$, we apply 
ECM. The detailed procedure is provided in the supplementary material(Section~S4), while a schematic flowchart is presented below.

\begin{algorithm}[ht] 
\caption{ECM Algorithm for Maximizing $\log L_{PC}^{(\text{full})}(\boldsymbol\Theta)$}
\KwIn{Observed data $\{\boldsymbol y_j,m_j\}_{j=1}^n$; initial $\boldsymbol\Theta^{(0)}$}
\KwOut{$\widehat{\boldsymbol\Theta}$}
\For{$t=0,1,2,\dots$ until convergence}{
  \tcp{E-step}
  Compute $\tau_{ij}^{(t)}$, $\delta_j^{(t)}$, $q_j^{(t)}$.

  \tcp{CM-step 1}
  Update $\boldsymbol\theta^{(t+1)}=\arg\max_{\boldsymbol\theta} Q_1(\boldsymbol\theta)$
  \;(\emph{quasi-Newton}).

  \tcp{CM-step 2}
  Update $(\boldsymbol\alpha^{(t+1)},\lambda^{(t+1)})=\arg\max_{\boldsymbol\alpha,\lambda} Q_2(\boldsymbol\alpha,\lambda)$
  \;(\emph{Newton/others; 1-D line search for $\lambda$}).
}
\end{algorithm}

\section{Approximation Validity}\label{AA}

\begin{figure}[htbp] 
    \centering
    \includegraphics[width=0.42\textwidth]{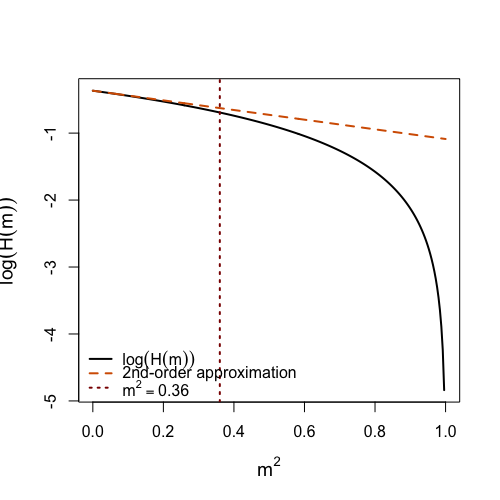}
    \caption{The solid shows the exact Shannon entropy, while the dashed denotes its
    second-order Taylor approximation. The vertical indicates $m^2 = 0.36$.}
    \label{fig:approx}
\end{figure}

Figure~\ref{fig:approx} shows the relationship between the logarithm entropy $\log(H(m))$ and the squared margin confidence $m^2$. The second-order Taylor approximation closely matches the true entropy function for $m^2 < 0.36$, indicating an approximately linear relationship in low-margin regions. 

Motivated by this observation, we adopt $m^2 = 0.36$ as a threshold. For samples with $m^2 \le 0.36$, which corresponds to posterior class probabilities in $[0.2, 0.8]$ and covers the majority of observations in practice, the squared margin confidence provides an effective and computationally efficient substitute for Shannon entropy when modeling the missingness mechanism.

\section{Simulations}

\subsection{Visualization of Simulations}
\begin{figure}[htbp]
    \centering
    \includegraphics[width=0.42\textwidth]{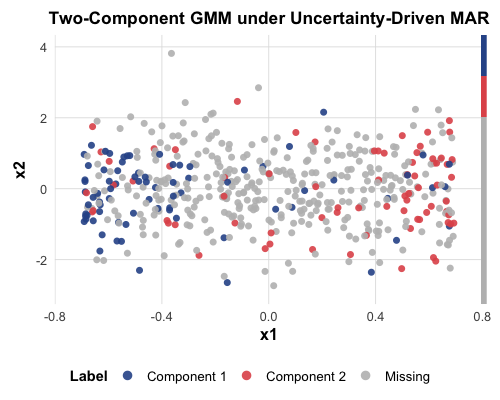}
    \caption{Blue and red points represent the two components, while grey points indicate missing labels. 
The right strip shows their proportions.
}
    \label{fig:r3}
\end{figure}

To study the MAR mechanism described in section~\ref{MM}, we conducted controlled simulations under a two-component Gaussian mixture model with equal variance matrix. The data generation process produced observations 
with overlapping densities, thus creating regions of high ambiguity around the classification boundary. After data generation, class labels were subject to deletion under MAR, calibrated to achieve an overall missing rate of 70\%.
The resulting dataset is visualized in Figure~\ref{fig:r3}, which shows the spatial distribution of labeled and unlabeled observations, as well as the overall proportion of missing labels.

\FloatBarrier

\subsection{Assessment of Simulations}
\begin{figure}[htbp]
    \centering
    \includegraphics[width=0.42\textwidth]{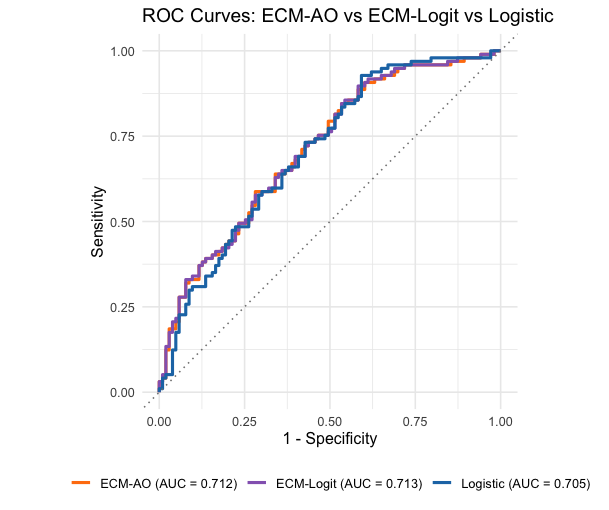}
    \caption{ROC curves comparing the three estimators under MAR.}
    \label{fig:roc_r4}
\end{figure}

To evaluate the impact of explicitly modeling the MAR mechanism on classification performance, three estimators were applied to the simulated dataset. Each method was assessed in terms of discriminative power, probabilistic calibration, and predictive performance. The quantitative results are presented in Table~\ref{tab:comp_metrics}, while Figure~\ref{fig:roc_r4} displays the associated ROC curves.

The results show that both ECM variants consistently outperform the logistic baseline. 
They achieve slightly higher AUC values (0.71 vs.\ 0.705) and substantially improved probabilistic calibration, with markedly lower LogLoss (0.67 vs.\ 0.97) and Brier scores (0.233 vs.\ 0.368).
At thresholds selected by Youden's index, both ECM variants attain F1-scores above 0.70, with the logit link favoring recall and the Aranda--Ordaz link yielding slightly higher precision. 
The added flexibility of the Aranda--Ordaz formulation may be advantageous in more complex missingness settings (Section~\ref{rob}).

\begin{table*}[htbp]
\centering
\caption{Comparative performance metrics of the three estimators under MAR missingness.}
\label{tab:comp_metrics}
\begin{tabular}{lcccccc}
\toprule
Method & AUC & LogLoss & Brier & Precision$_{\text{opt}}$ & Recall$_{\text{opt}}$ & F1$_{\text{opt}}$ \\
\midrule
ECM (AO link)        & 0.712 & 0.675 & 0.233 & 0.599 & 0.845 & 0.701 \\
ECM (Logit link)     & 0.713 & 0.673 & 0.233 & 0.592 & 0.897 & 0.713 \\
Logistic baseline    & 0.705 & 0.971 & 0.368 & 0.143 & 0.072 & 0.096 \\
\bottomrule
\end{tabular}
\end{table*}

\begin{figure}[htbp]
    \centering
    \includegraphics[width=0.42\textwidth]{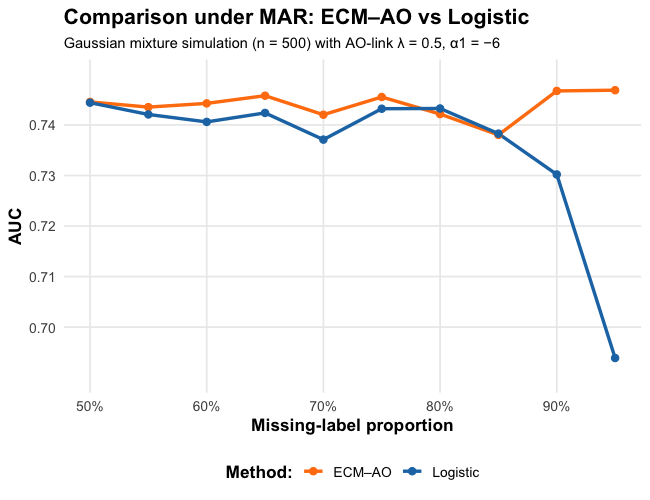}
    \caption{The AUCs of two estimators under different missing proportion}
    \label{fig:roc_r8}
\end{figure}

In addition, we compare the performance of the ECM--AO and logistic across different missing-label proportions.
Figure~\ref{fig:roc_r8} shows a clear trend in increasing proportions of missing-labels. 
The ECM--AO maintains stable discrimination, whereas logistic regression progressively degrades. 
This contrast reflects the MAR mechanism, under which training on observed labels alone induces selection bias, while explicitly modeling missingness allows the ECM--AO to preserve discrimination and calibration as labeled information diminishes.

\subsection{Verification of Algorithmic Robustness}\label{rob}

To rigorously evaluate robustness under distributional misspecification, we extend the analysis beyond the Gaussian assumption and assess the proposed ECM--AO estimator under three non-Gaussian mixture settings: Gamma, Beta and Laplace mixtures. Together, these settings violate key Gaussian assumptions, including symmetry, infinite support, and tail behavior\cite{Webb2000,gupta2004,song2014}. This provides a stringent test of robustness.

To implement this test, we evaluated the ECM--AO estimator on three two-dimensional non-Gaussian mixture settings and compared it with a logistic regression baseline that ignores the missingness mechanism. 
As shown in Table~\ref{tab:comp_mar_all}, ECM--AO performs comparably or better across all settings. 
Although AUC values remain stable across mixtures, ECM--AO achieves substantially lower LogLoss and Brier scores and maintains a favorable precision--recall trade-off, with higher recall and no notable loss of precision. 
In contrast, logistic regression exhibits degraded calibration and recall under MAR. 
These results indicate that the proposed framework remains robust under moderate distributional misspecification.

\begin{table*}[htbp]
\centering
\caption{Comparative performance metrics under MAR missingness across different generative mixtures.}
\label{tab:comp_mar_all}
\begin{tabular}{llccccc}
\toprule
Distribution & Method & AUC & LogLoss & Brier & Precision$_{\text{opt}}$ & Recall$_{\text{opt}}$ \\
\midrule
\addlinespace[2pt]
\textbf{2D Gamma mixture} & ECM (AO link)        & 0.732 & 0.862 & 0.244 & 0.664 & 0.778 \\
                          & Logistic baseline    & 0.744 & 1.173 & 0.431 & 0.350 & 0.485 \\
\addlinespace[2pt]
\textbf{2D Beta mixture}  & ECM (AO link)        & 0.668 & 0.683 & 0.241 & 0.609 & 0.857 \\
                          & Logistic baseline    & 0.689 & 1.107 & 0.402 & 0.257 & 0.184 \\
\addlinespace[2pt]
\textbf{2D Laplace mixture} & ECM (AO link)      & 0.696 & 0.677 & 0.234 & 0.670 & 0.702 \\
                            & Logistic baseline  & 0.686 & 1.006 & 0.377 & 0.365 & 0.337 \\
\bottomrule
\end{tabular}
\end{table*}

\section{MAGIC Gamma Telescope Dataset}
\subsection{Data Description}
The empirical analysis in this study is based on the MAGIC Gamma Telescope dataset obtained from the UCI Machine Learning Repository. The dataset comprises 19,020 observations. Each observation is described by ten continuous image variables, together with a binary class label indicating the true particle type (\textit{gamma} or \textit{hadron}).

\begin{figure}[htbp]
    \centering
    \includegraphics[width=0.42\textwidth]{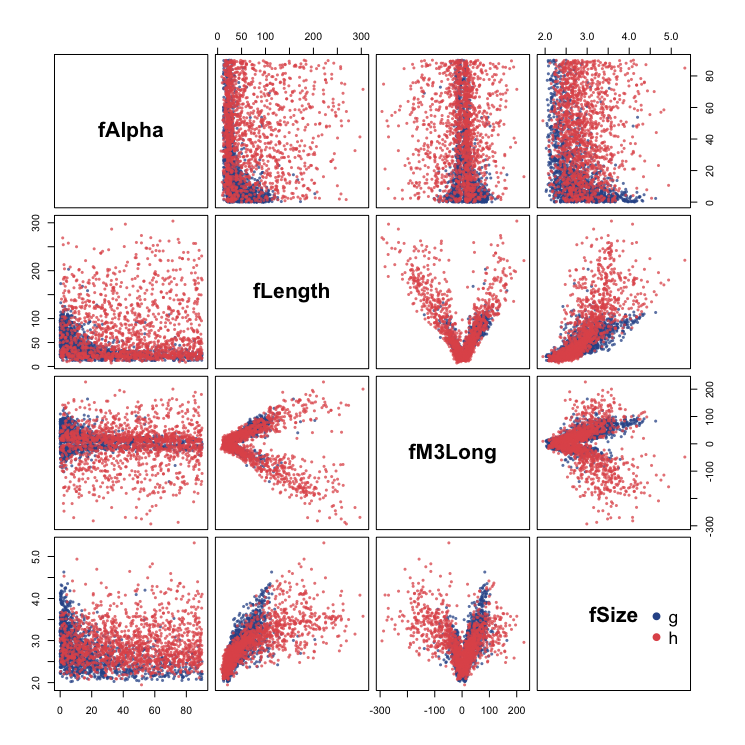}
    \caption{Pairwise scatterplot matrix of the four variables from MAGIC dataset. 
    Blue and orange points represent \textit{gamma} and \textit{hadron} events, respectively. }
    \label{fig:magic_pairs}
\end{figure}

Following the data preprocessing procedure, four representative variables, \textit{fAlpha}, \textit{fLength}, \textit{fM3Long}, and \textit{fSize}, are retained for subsequent modeling.
Figure~\ref{fig:magic_pairs} then shows that the pairwise scatterplot matrix exhibits substantial overlap between the two particle classes, suggesting that a simple linear boundary would be insufficient for discrimination.

Building on the dataset above, we construct a controlled missingness mechanism under the MAR assumption.
This procedure results in approximately 70\% of the labels missing, producing a partially labeled MAGIC dataset for empirical evaluation.

\subsection{Model Comparison}

To further evaluate the classification performance of the ECM--AO, we compare it with logistic regression under different classification thresholds. 

\begin{figure}[htbp]
    \centering
    \includegraphics[width=0.42\textwidth]{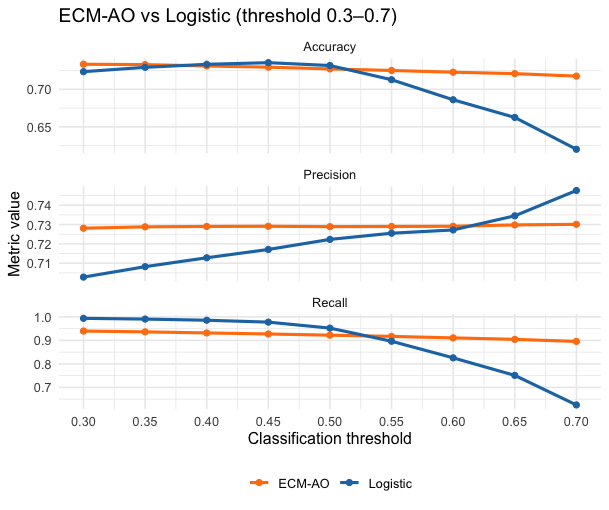}
    \caption{Precision, Recall, and Accuracy across classification thresholds (0.3–0.7) for ECM–AO and logistic under 70\% label missing.}
    \label{fig:threshold_metrics}
\end{figure}

Figure~\ref{fig:threshold_metrics} shows performance as the classification threshold varies from 0.3 to 0.7. Across this range, ECM-AO maintains higher accuracy and more stable recall and precision than logistic. In contrast, the logistic model exhibits a marked drop in accuracy and recall as the threshold increases, indicating greater sensitivity to the cutoff choice. This stability suggests that ECM-AO produces smoother and more robust decision boundaries across thresholds.

\subsection{Effect of Missing Proportion on Model Performance}
\begin{figure}[htbp]
    \centering
    \includegraphics[width=0.42\textwidth]{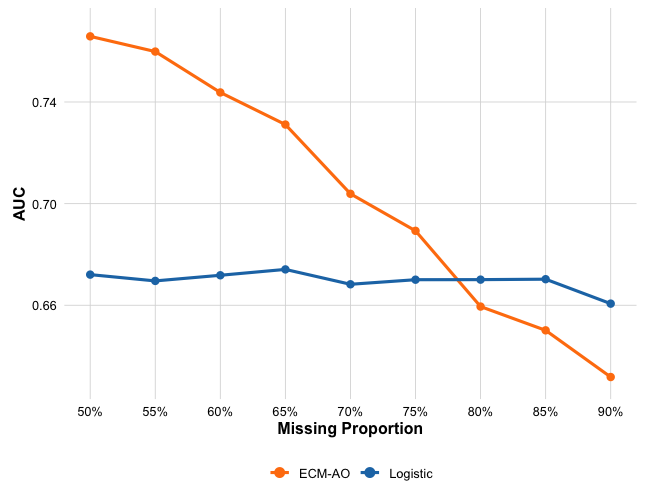}
    \caption{Comparison of AUC between ECM--AO and logistic under varying levels of missing proportion. 
Points denote mean AUC over multiple realizations.
}
    \label{fig:missing_proportion}
\end{figure}

As shown in Figure~\ref{fig:missing_proportion}, we compare ECM--AO with logistic regression under increasing levels of missing proportion (50--90\%).
For each missing level, multiple datasets are generated under the MAR mechanism, and performance is summarized by the mean AUC across repeated runs. 
As expected, classification performance degrades for both models as the missing rate increases. 
ECM--AO consistently outperforms the logistic baseline at moderate missing levels (50\%--70\%). 
However, the performance gap narrows as missingness increases and reverses when the missing rate approaches 90\%. 
This behavior is consistent with the reduced information of the MAR-based uncertainty modeling under extreme label sparsity, where the advantage of the adaptive AO link becomes less pronounced.

\section{Conclusions}

This study demonstrates that explicitly modeling label missingness within a semi-supervised framework leads to more robust and better-calibrated classification under MAR. 
Across both simulated and real data, the proposed ECM--AO consistently outperforms the logistic baseline that ignores the missingness mechanism. 
It achieves improved calibration while maintaining stable discrimination under moderate label missingness. 
The additional flexibility of the AO link enables a smoother mapping from uncertainty to missing probability, yielding more reliable posterior estimates. 
Although the advantage diminishes under extreme missingness, 
ECM--AO remains numerically stable and competitive.

A remaining challenge is extending the framework to the mixture model with more than two components, as the current margin confidence relies on the two largest posterior probabilities. 
Developing uncertainty measures that exploit the full posterior vector is a promising direction to extend to more complex mixture settings.

\section{Supplementary}
Further theoretical derivations and algorithm details are provided in the supplementary material (Sections~S1--S5), available at:
\begin{center}
\url{https://github.com/LeomusUNSW/IJCNN}
\end{center}

\bibliographystyle{IEEEtran}  
\bibliography{references}

\end{document}